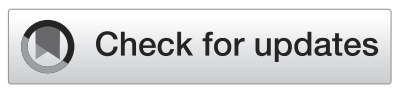

*CORRESPONDENCE
Ali A. Safieh,
alisafia955@gmail.com

SPECIALTY SECTION
This article was submitted to Computational Intelligence in Robotics, a section of the journal Frontiers in Robotics and AI

# End-to-end Jordanian dialect speech-to-text self-supervised learning framework

Ali A. Safieh*, Ibrahim Abu Alhaol and Rawan Ghnemat

Data Science Department, King Hussein School of Computing Sciences, Princess Sumaya University for Technology, Amman, Jordan

Speech-to-text engines are extremely needed nowadays for different applications, representing an essential enabler in human–robot interaction. Still, some languages suffer from the lack of labeled speech data, especially in the Arabic dialects or any low-resource languages. The need for a self-supervised training process and self-training using noisy training is proven to be one of the up-and-coming feasible solutions. This article proposes an end-to-end, transformers-based model with a framework for low-resource languages. In addition, the framework incorporates customized audio-to-text processing algorithms to achieve a highly efficient Jordanian Arabic dialect speech-to-text system. The proposed framework enables ingesting data from many sources, making the ground truth from external sources possible by speeding up the manual annotation process. The framework allows the training process using noisy student training and self-supervised learning to utilize the unlabeled data in both pre- and post-training stages and incorporate multiple types of data augmentation. The proposed self-training approach outperforms the fine-tuned Wav2Vec model by 5% in terms of word error rate reduction. The outcome of this work provides the research community with a Jordanian-spoken data set along with an end-to-end approach to deal with low-resource languages. This is done by utilizing the power of the pretraining, post-training, and injecting noisy labeled and augmented data with minimal human intervention. It enables the development of new applications in the field of Arabic language speech-to-text area like the question-answering systems and intelligent control systems, and it will add human-like perception and hearing sensors to intelligent robots.

KEYWORDS

**transformers, Wav2Vec, self-supervised, speech-to-text, self-training, HCI, robotics, embedded system**

## 1 Introduction

Speech is one of the most effective and important human–computer interaction methods. Through text-to-speech and speech-to-text functionalities, humans can interact easily with digital systems that enables humans to achieve tasks efficiently (Karray et al. (2008). Automatic speech recognition (ASR) is the process of decoding voice signal sequences (spectrogram) into a sequence of phonemes that will build a sequence of words on a function of time. Although the ASR or "Speech to text" functionality can be considered as a bottleneck for several types of applications such as the informative and actionable systems, it remains to be the core prerequisite of control applications and conversational question-answering systems. The complexity of the problem is derived from the nature of the data, in addition to the transformation and modeling aspects. Data in the form of speech require special treatment and modeling since these comes in a stream, so models with poor contextual representations may fail in handling such contextual data. Some implementations require simple models, especially in the control applications and Internet of Things (IoT) solutions since they are based on limited vocabulary. On the other hand, advanced applications and systems depend on continuous speech recognition in an offline and online manner. In fact, natural language processing and text mining are essential subsystems to solve complex problems such as customer feedback analysis, conversational chatbot, and topic/emotion recognition. Our methodology is to provide a comprehensive end-to-end approach for handling low-resource languages such as the Arabic dialect by utilizing the unlabeled data in the pretraining and post-training using self-supervised training and noisy student training. We could outperform the base Wav2Vec by 5% word error rate (WER) by applying a self-training approach on our obtained data set and by providing several options for deploying the provided model in real-life applications such as embedded systems, robotic, or question-answering systems.

# 2 Literature overview

## 2.1 Jordanian Arabic speech

Jordanian Arabic (JA) is the primary language spoken in the Hashemite Kingdom of Jordan, and it consists of various dialects. The derived dialects can be further classified into three main categories: (i) urban, (ii) rural, and (iii) bedouin. The rural dialect comes mainly from the Jordanian Hauran, which is an extension of the Syrian Hauran Plain Sakarna (1999). JA is one of the richest and most complex languages due to its phonological and morphological nature, since it has more additional phonemes than the Modern Standard Arabic (MSA) phoneme set. These dialects were affected by successive migrations of Arab workers and displaced persons and their entry into the Jordanian society. **Table 1** provides some examples that show the differences between the phonological representation of the same word in different Jordanian dialects.

**TABLE 1** **Jordanian dialects diversity.**

| Standard | Rural/bedouin | Urban | Rural | Meaning |
|---|---|---|---|---|
| qatal | Gatal | ‘alal | katal | killed |
| kalb | Tshalb | kalb | tshalb | dog |
| kaif | tshef | kef | tshef | how |
| thaq | d ag | da’ | dhag | he tasted |

## 2.2 ASR system history

**Figure 1** shows the main components of the traditional automatic speech recognition systems. Most of these systems or pipelines consist of common steps such as preprocessing, modeling, results representation, and an evaluation step Karpagavalli and Chandra (2016).

### 2.2.1 Preprocessing step

The Auditory Front End is one of the common techniques used in the preprocessing step and it works by obtaining the signal of the voice and converting the signal into an auditory-based representation or in other words into a speech vector. An example of this is the Mel-Frequency Cepstral Coefficients (MFCC) Han et al. (2006).

### 2.2.2 Acoustic model

After the feature extraction step, the speech frame is passed as an input vector to the modeling layer that can be considered as a classification layer Karpagavalli and Chandra (2016). This will try to map the input frame into a sequence of phonemes and search for the optimal word representation based on the language model. This model can be a neural-based or simple statistical model, for example, the Gaussian mixture model (GMM), artificial neural network (ANN), and support vector machine (SVM).

### 2.2.3 Language model

The statistical language model is an N-gram–based model that will produce the probability of the word given the word (n − 1), and this N-gram model can also detect multi-word terms and consider the context of the word. The language model incorporates the acoustic model to enable the search for the optimal sequence of words. The language model can be a simple probabilistic N-gram model or neural and attention-based model Jelinek et al. (1991).

### 2.2.4 Model metrics and evaluation

The evaluation process of the ASR systems is similar to the evaluation of the sequence labeling systems such as machine translation. The selected evaluation metric in most of the production systems is the word error rate (WER) metric Amodei et al. (2016) Baevski et al. (2020) since it will take into account the position, absence, and the additional words between the predicted and target sentences, in addition to measuring the efficiency of the language model.

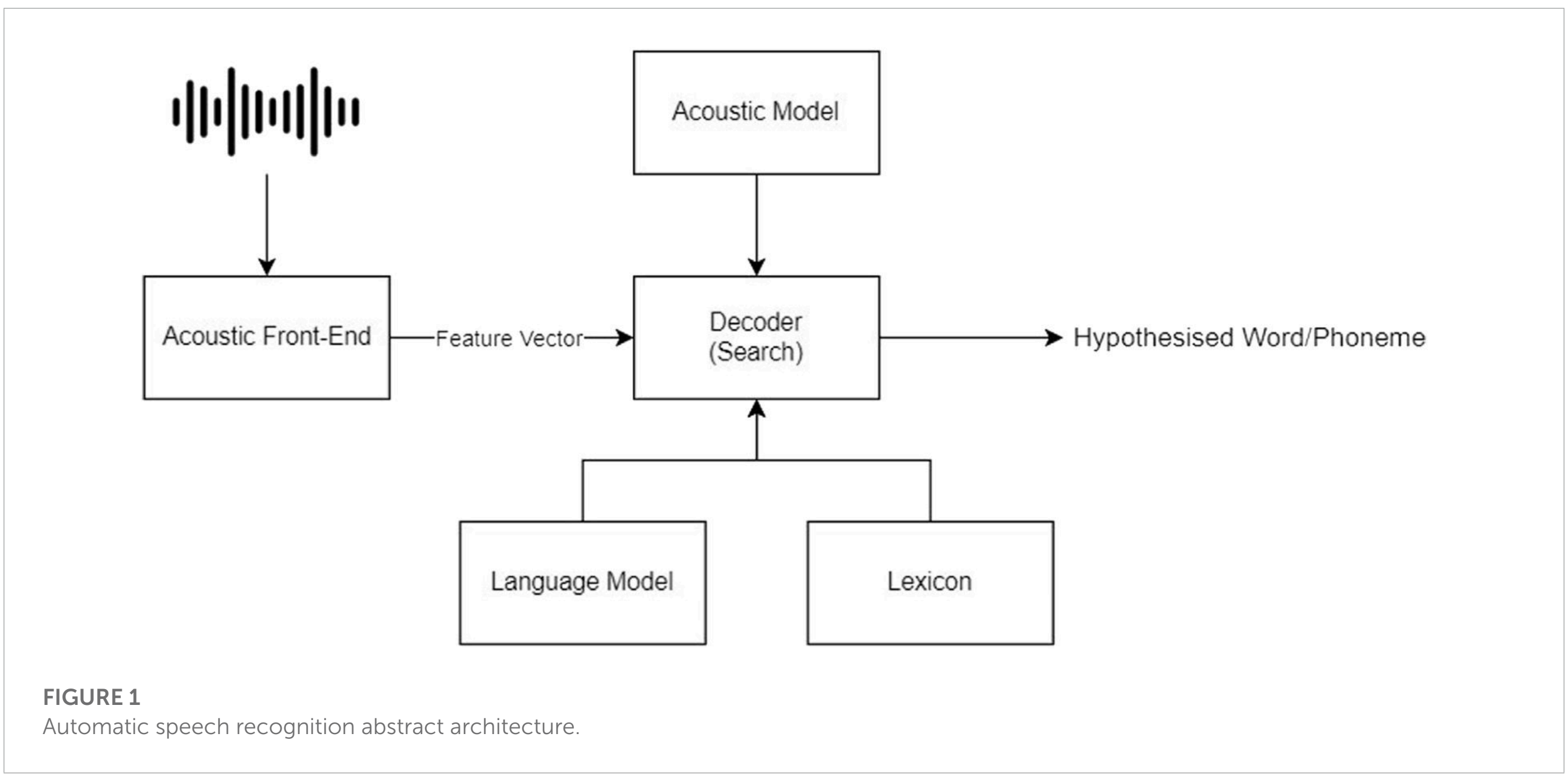

FIGURE 1
Automatic speech recognition abstract architecture.

#### 2.2.4.1 Word error rate metric

The WER metric is defined as the minimal cost of converting string 1 into string 2, using three operations (substitution, insertion, and deletion). The WER is one of the most famous metrics to evaluate speech recognition engines. It is calculated as the edit distance, divided by the number of words in the reference string.

$$WER = \frac{(S + D + I)}{N} \tag{1}$$

where S is the number of substitutions, D is the number of deletions, I is the number of insertions, and N is the number of words in the reference.

#### 2.2.4.2 Character error rate metric

The character error rate (CER) is similar to the WER. It measures the number of substitutions, deletions, and insertions to convert the predicted sequence into the target MacKenzie and Soukoreff (2002). It is used to analyze character level errors and optimize the alignment between the predicted and target texts.

## 3 Methodologies and paradigms

The ASR systems are classified based on multiple dimensions. One of the dimensions that we are interested in is the "style of learning." In the machine learning world, there are two main styles of learning: end-to-end (E2E) and the multistage modeling.

**Figure 2** shows the traditional multilingual model can have a dedicated language model for each language and that will fail in the scarce languages. By introducing the E2E model we can utilize data from different languages and dialects Kannan (2022). That will improve the scarce languages and decrease the level of customization as shown in the purple blocks in **Figure 2**. The case in Jordanian Arabic, which is considered a scarce language, is similar to the proposed solution in Kannan (2022), therefore we can utilize the pretrained models and data sets, especially the Modern Standard Arabic data sets and models.

### 3.1 End-to-end models

The E2E models are the most effective and recent methodologies that handle complex problems with deep learning models. The power of E2E models is derived from the data complexity and assumptions. It does not require having deep business knowledge about the problem and the data set. As a result, it can efficiently encapsulate the mapping between the input and target in one single deep learning model. E2E models are very efficient in solving problems with very minimal supervision and data transformation. This is especially true in natural language processing, which includes part of speech taggers, text classification, and named entity recognition. This will help reduce the effort of feature engineering and understanding. E2E models can be efficiently optimized by replacing the entire pipelines of feature engineering and voice frame alignment with neural-based models. Not only that, but it also allows us to handle noisy environments, different dialects, and different languages Yalta (2020).

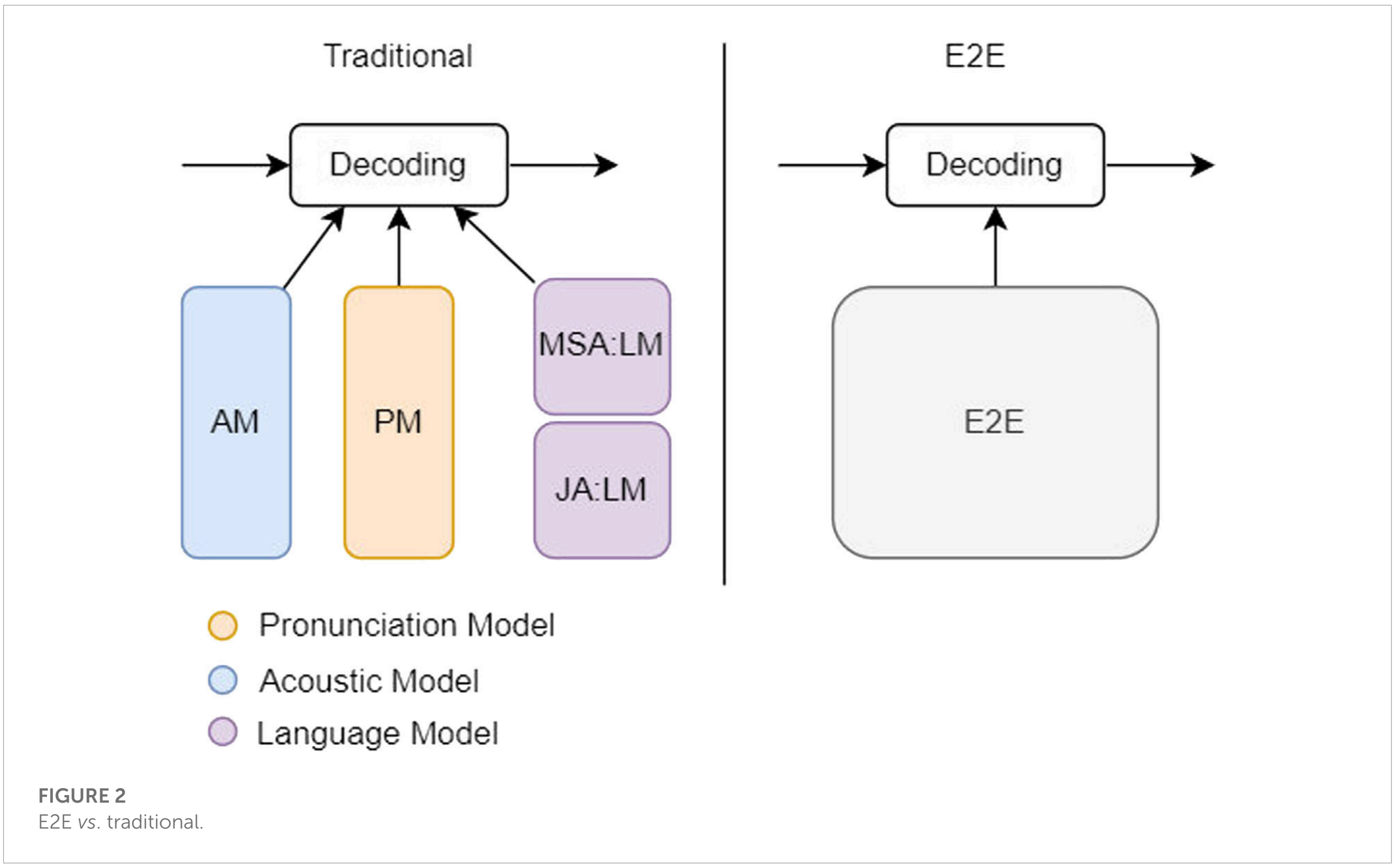

FIGURE 2
E2E *vs*. traditional.

### 3.1.1 End-to-end limitations

Similar to all deep learning models, the E2E models suffer from a lack of interpretability, predictability, and diagnosability. Moreover, the E2E models cannot be modified after training, and the state-of-the-art models cannot be integrated with it as a subtask of the whole training process since it will not be considered as E2E anymore. The main obstacle of the E2E approach is data intensive, so it requires a huge amount of data to complete the training task which is not feasible in most cases.

## 3.2 Multistage Learning

In the 1980s, after IBM applied the hidden Markov model (HMM) in ASR, multistage learning was the most commonly used approach in the previous decades, especially in the limited vocab ASRs. The traditional learning approach is the straightforward way to build an ASR since it breaks down the entire process into subtasks which include understanding the voice, transforming it into phonetic form, and building the language model. In most cases, the multistage model is a statistical model that can be traced easily and requires less computing power. It can also achieve suitable results in most of the limited vocabulary and context-independent ASRs Fendji et al. (2022).

### 3.2.1 Traditional learning limitations

On the other hand, traditional ASR systems require more feature engineering tasks. They also require the voice to be accurately aligned with the transcript Yalta (2020). In some cases, the data are required to be aligned in the frame level, except in the Connectionist Temporal Classification (CTC), as it requires only a language model.

# 4 Available tools and data sets

This section will review previous work which includes the available data sets and open-source tools and methodologies.

## 4.1 Speech recognition open source tools

Kaldi and CMUSphinx are the most widely used in the previous decades. Both of them show a good performance in multistage ASRs. As described in Povey et al. (2011), Kaldi can outperform CMUSphinx in terms of WER. Also, both of them require dedicated language model training. Recently, Wav2Vec 2.0 and DeepSpeech are the most popular models in this area. Thus, they can achieve the best performance in comparison with the other model. Wav2Vec and DeepSpeech are considered E2E models and both of them do not require additional featureengineering and language model building.

TABLE 2 ASR open source tools.

| Features | Toolkits | | | |
|---|---|---|---|---|
| Name | Kaldi Povey et al. (2011) | CMUSphinx Lamere et al. (2003) | Facebook Wav2Vec Baevski et al. (2020) | DeepSpeech Amodei et al. (2016) |
| Paradigm | Multistage model | Multistage model | E2E | E2E |
| Algorithm | Gaussian mixture model | Hidden Markov model and Gaussian mixture model | Self Supervised Learning | RNN |
| Language | C++ | Java | Python | Python |
| APIs | C++ and Python | Python and Java | Python | Python |
| GPU Training | Not available | Not available | Available | Available |

TABLE 3 ASR datasets.

| Features | Datasets | | |
|---|---|---|---|
| Name | Arabic Speech Corpus Halabi (2016) | Arabic Common Voice Corpus moz (2022) | MASC: Massive Arabic Speech Corpus Al-Fetyani et al. (2021) |
| Duration | 3.7Hr | 137Hr | 1000Hr |
| Accent | South Levantine Arabic | MSA and Dialectal | Multi-dialect |
| Link | Arabic Speech Corpus | Arabic Common Voice Corpus | MASC |

Wav2Vec is trained on unlabeled data to generate the best feature representation that can feed the downstream task. As a result, Wav2Vec requires less data for convergence, and it is good for scarce languages. In Yi et al. (2020), Wav2Vec achieved state-of-the-art results and outperformed DeepSpeech in the ASR task. Table 2 provides a comparison between the popular speech recognition toolkits in terms of algorithm, programming language APIs, and GPU training capability.

## 4.2 Public data sets

This section is to identify the famous Arabic spoken data sets. The Arabic Speech Corpus Halabi (2016) is a high-quality Arabic speech corpus recorded in a Damascian accent. This voice data set is perfectly aligned with the transcript and optimized for the text-to-speech task. It has additional time boundaries for the phonemes. Also, the provided transcript is an orthographic transcript in the Buckwalter format. Arabic Common Voice Corpus moz (2022) is a human-annotated speech data set. It is an MSA and multi-dialect data set created by Mozilla's common voice project. Massive Arabic Speech Corpus (MASC) Al-Fetyani et al. (2021) is a multiregional, multi-genre, and multi-dialect data set. It is the largest annotated Arabic data set. Table 3 provides the available Arabic annotated datasets along with the duration and spoken accent.

# 5 Integration with embedded systems

Integrating an automatic speech recognition engine with edge devices and robots can be considered the primary bottleneck, due to the difficulties in integrating the SOTA DNN-based models with low computing power and power consumption. To tackle this problem, there are multiple integration styles that can be categorized into online and offline integrations. The online speech recognition integration can be designed as a streaming API between a high-performance centralized server that can support multiple devices. In the online integration, there is no need to consider the delay and power consumption since no heavy computations will be done on the device side, and you can easily scale up the centralized server to speed up serving the other system. On the other hand, the online approach requires a permanent internet connection between the devices. Also, the speech recognition server can be considered a single point of failure.

The offline integration provides stable integration of the speech recognition service on the same edge device such as a robot, smartphone, or smartwatch without requiring an internet connection.

The integration of speech recognition can be done using multiple approaches on the Raspberry Pi and NVIDIA Jetson Nano for robots, mobile phones, or any smart device Gondi (2022).

The offline consumption of the deep learning STT models can be done on both CPU or GPU on the Jetson Nano, by deploying the model on TorchScript. TorchScript can optimize the neural-based speech recognition models such as Wav2Vec and deep speech and convert them into intermediate formats, and then optimize and compress the model using quantization. This reduces the model size and optimizes the mathematical operations on the CPU or GPU Gondi and Pratap (2021).

# 6 Data set

We have a shortage of labeled spoken data in the Arabic language, especially in the Jordanian dialect. Therefore, enriching the existing data sets is mandatory. In this thesis, we aim at utilizing public videos and audio files on the social media platform, especially on YouTube and Metaverse, to enrich the variable data set.

## 6.1 Data set sources

To ensure the diversity of the obtained accents and topics, the collected data set was ingested from multiple channels, and it includes recordings from internet influencers, street interviews, news, and political discussion to represent a wide range of dialects.

## 6.2 Data transformation and augmentation

The data cleansing process is an important step before conducting the labeling sessions to convert the data set into an easily manageable format. Multiple data transformations and cleansing techniques were applied to the data set which included voice and text labels. First of all, splitting the data set into small files on the basis of silence using voice activity detection (VAD) wiseman (2022) is a crucial step to make the recordings easily manageable and annotated. The whole data were downsampled to 16 kHz, and all records with a duration more than 20 s and lower than 1 s were excluded. Moreover, the text-to-speech ratio was checked for all records to exclude and avoid any misalignment. From the text side, special characters, Arabic diacritics, punctuations, and repeating characters were removed from the text, in addition to normalizing some characters that will lead to the same phonemes.

As part of the noisy student training, noisy data batches were injected at every generation to ensure robustness and generalization of the model, and to decrease the dependency of the environment on the model Zhang et al. (2020).

Three types of augmentations were applied to each data generation. Gaussian noise is well known as the additive noise method used to ensure the generalization of the model by adding Gaussian distributed noise Braun et al. (2017). Furthermore, time stretch and pitch shift were also applied to make the model robust to environmental effects and against misalignment issues Schlüter and Grill (2015).

## 6.3 Labeling process

### 6.3.1 Google cloud speech recognition

The labeling process is a complicated task. The Google speech API can handle the initial labeling process since it is one of the most robust and efficient solutions in the market, and it can handle the Jordanian dialect. The generated transcripts from the google speech API is used to initially label the data set to facilitate the entire process on the annotator side.

### 6.3.2 Manual corrections

This research was conducted on the Jordanian dialect and, at some point, slang language. Therefore, we are required to have more accurate data than to only fetch the labels from Google, and not completely rely on Google engine accuracy. For this reason, we started annotation sessions to maximize data accuracy by distributing the annotation tool. To speed up the annotation process, we deployed a full speech-to-text annotation tool called Doccano Nakayama et al. (2018). It is a powerful and collaborative tool that can handle the cooperation between annotators and approvers.

## 6.4 Data set characteristics

The obtained data set consisted of 97,000 records for multiple speakers, and the overall duration of the spoken data set was 113 h, of which 52 h were labeled using Google Cloud Speech and manual transcript.

### 6.4.1 Topics and term frequency

The whole corpus contained around 35,000 terms without stemming or lemmatization and without removing stop words. The recordings revolved around different topics and different domains such as political, tourism, and religious discussions. **Figure 3** demonstrates the term frequencies of the obtained corpus.

### 6.4.2 Records and voice characteristics

The dataset consists of 97 thousand records 37 thousand of them are labeled, with mean recordings duration 4.74 s,

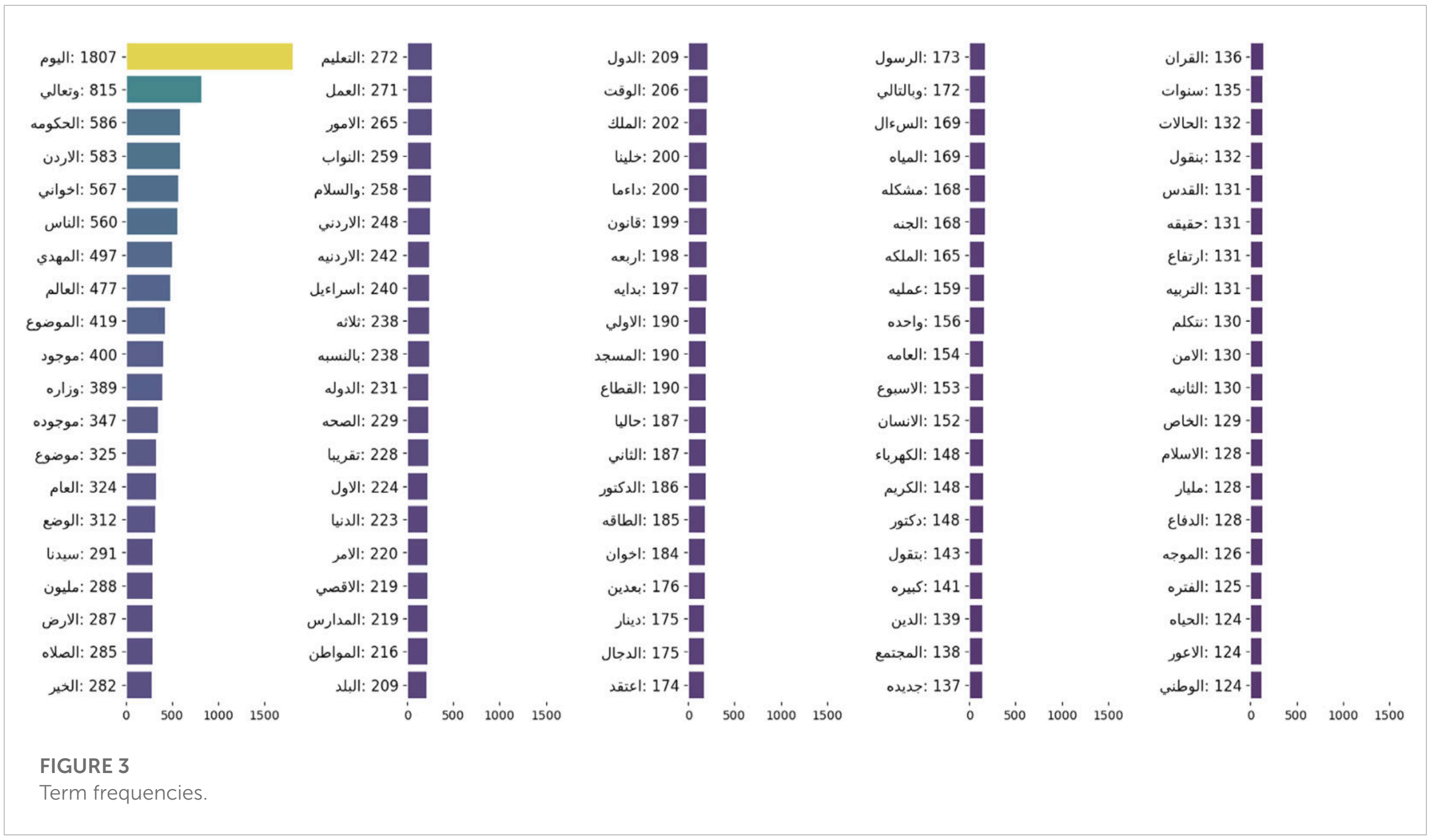

FIGURE 3
Term frequencies.

in **Figure 4** demonstrates the distribution of the recording length.

# 7 Speech-to-text modeling

This section provides an overview of self-supervised learning (SSL), self-training (noisy student training), and Wav2Vec which were intended to be used in our research.

## 7.1 Self-supervised learning

Self-supervised learning (SSL) is an efficient technique to leverage and detect the underlying structure of the unlabeled data based on the occurrences of the instances and turn it into the predictive capability to predict the unobserved instances Zhai et al. (2019). SSL is one of the most recent methodologies that can obtain state-of-the-art results, especially in one- or few-shot learning. This is because it can be trained on huge unlabeled data sets to exploit them and solve downstream tasks. Because of the lack of labeled data sets in most of the problems, SSL can utilize the unlabeled data to build a robust representation system. This can be used as a generalized vectorization model for another downstream task, instead of building a model on fewer labeled instances Chaudhary (2022). **Figure 5** demonstrates applying self-supervised learning in natural language processing.

## 7.2 Wav2Vec 2.0

The most recently published state-of-the-art model in the speech representation and recognition field is the Metaverse Wav2Vec model, which is a self-supervised contrastive model. Also, the Wav2Vec model can achieve the best results in comparison with the other approaches, especially on small data sets. The Wav2Vec model consists of two stages Baevski et al. (2020) as shown in **Figure 6**.

### 7.2.1 Self-supervised stage

This contrastive task is to utilize the unlabeled data set to build a meaningful representation layer as an embedding layer. This is a crucial task that enables outperforming the other models even on a small data set.

### 7.2.2 Latent representation layer

This layer consists of multiple temporal convolutional units that take the raw data input and convert it into latent representation, a layer of normalization, and GELU activation function Kessler et al. (2021).

### 7.2.3 Quantization

This layer is responsible for combining and discretizing the latent representation into a finite number of values or phonemes. This can be achieved using the Gumbel Softmax equation Eq. 2 as a classification layer to classify and combine multiple latent representations that represent the same phoneme into the same codeword from the codebook.

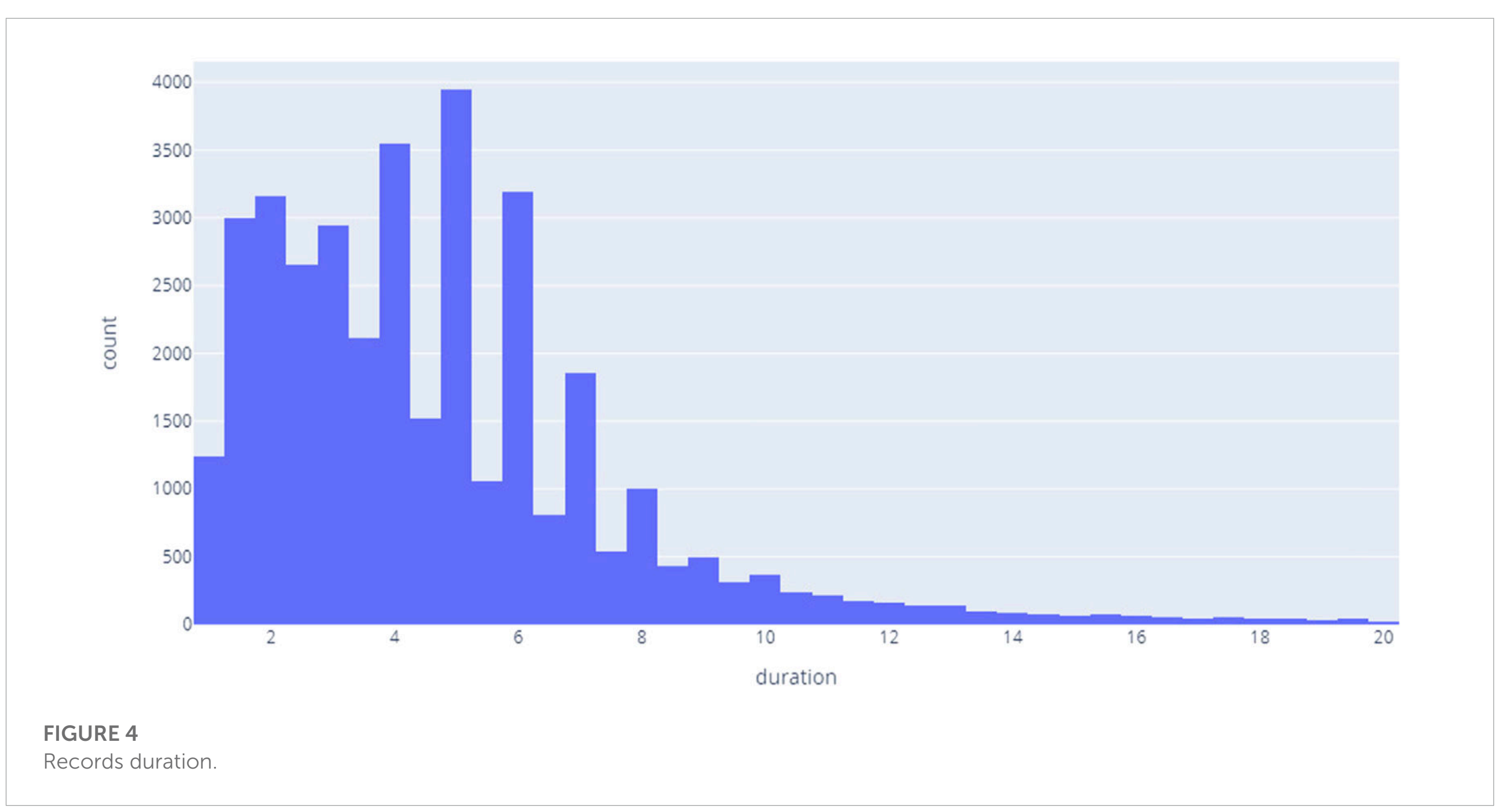

FIGURE 4
Records duration.

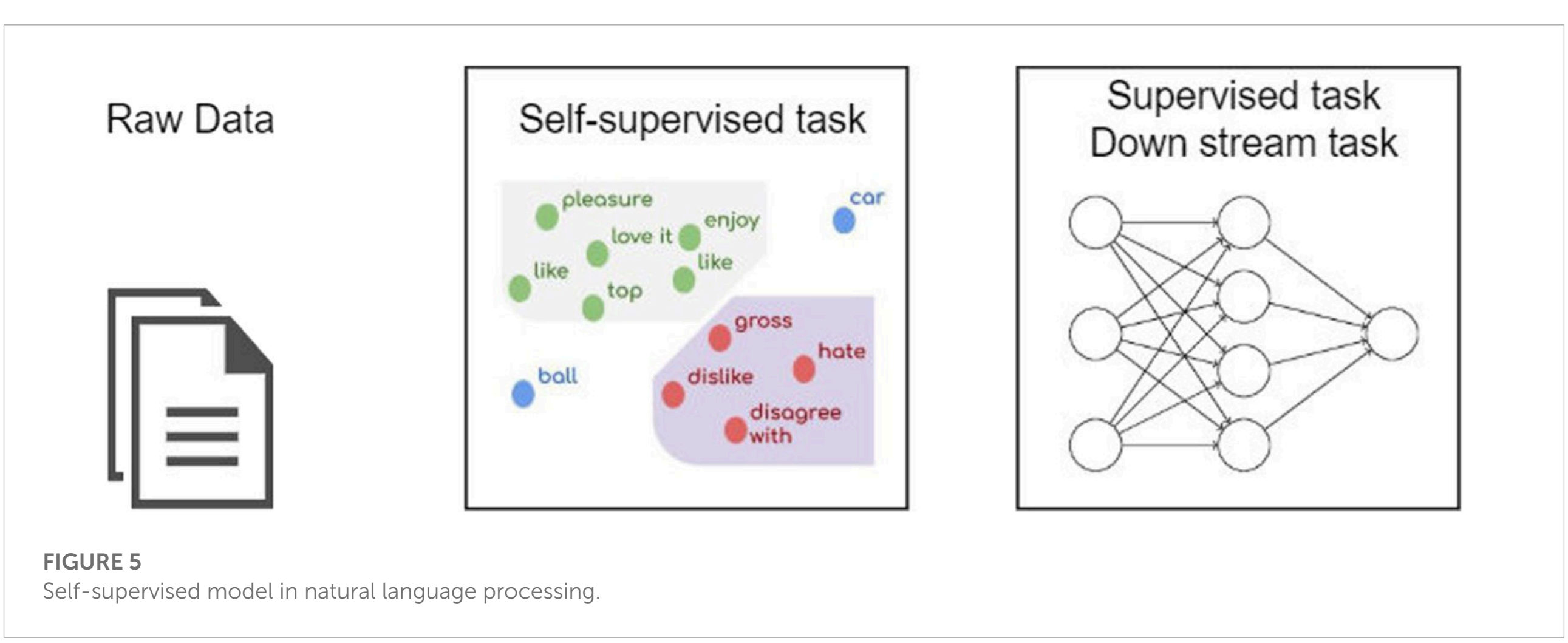

FIGURE 5
Self-supervised model in natural language processing.

$$p_{g,v} = \frac{exp\left(sim\left(l_{g,v} + n_v\right)\right)/\tau)\right)}{\sum exp\left(\left(l_{g,v} + n_v\right)/\tau\right)\right)} \quad (2)$$

where sim is cosine similarity,

l is logits calculated from z,

n is –log (–log(u)), and

u is sampled from the uniform distribution U(0, 1).

*τtemperature*.

### 7.2.4 Contextualized feature representation

After encoding the input raw data into a latent representation, the output of the temporal convolutional block is fed into a deep transformer layer that will take the sequence and context into consideration Baevski et al. (2020).

### 7.2.5 Fine-tuning stage

The adaptation of the Wav2Vec speech representation model to serve the ASR task in Yi et al. (2020) shows remarkable improvements over RNN-LSTM and speech transformers for low language resources. The fine-tuning process is done by adding a randomly initialized linear projection layer on top of the Wav2Vec context network and freezing the feature extractor. This layer can act as a classification layer and each class C represents vocabulary Kessler et al. (2021).

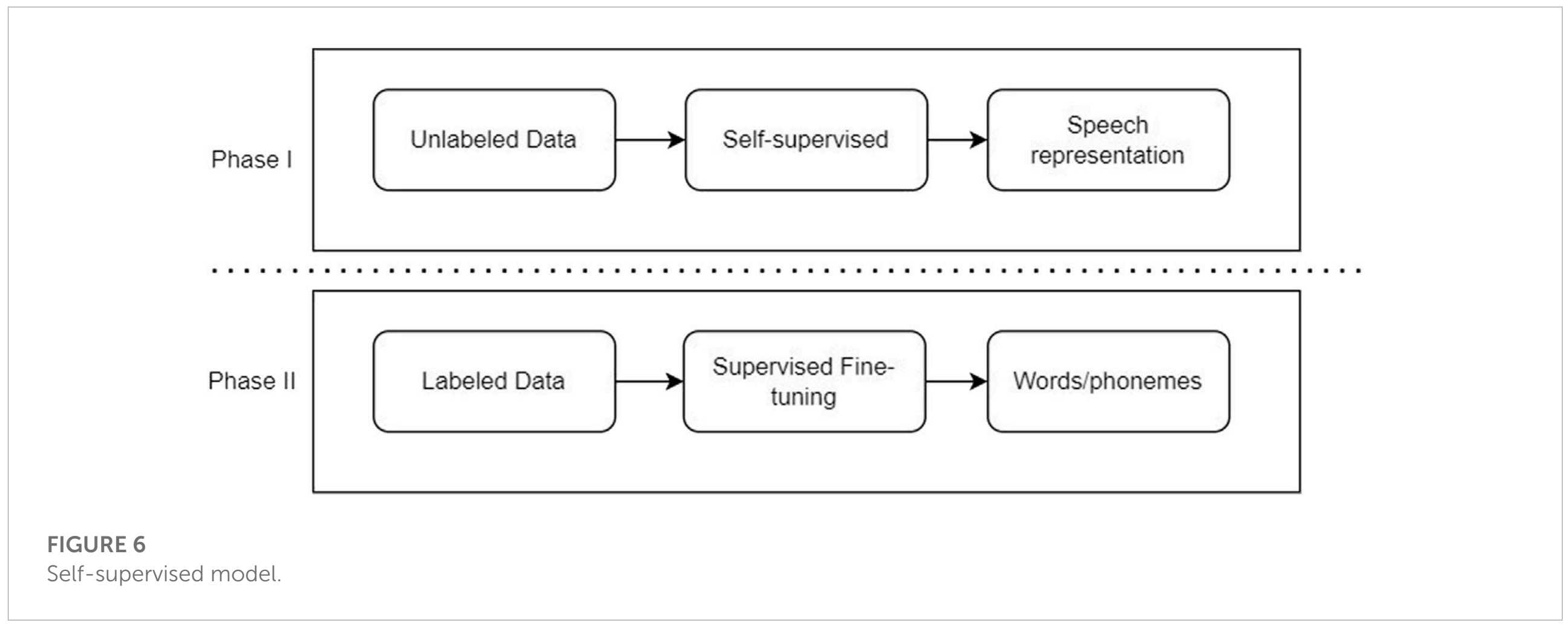

FIGURE 6
Self-supervised model.

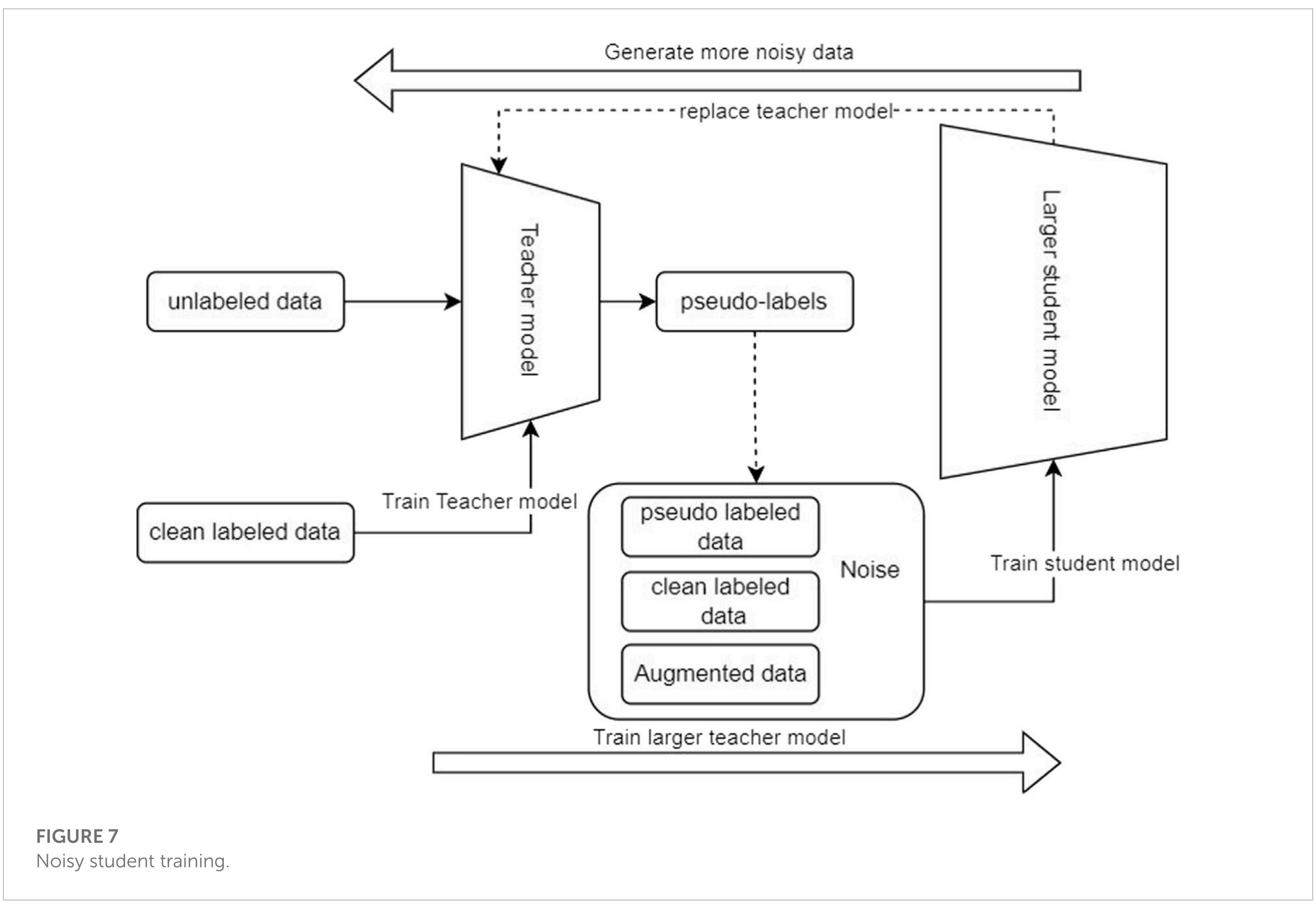

FIGURE 7
Noisy student training.

## 7.3 Self-training (noisy student training)

Self-training is a method in the semi-supervised training paradigm that has shown remarkable enhancement in the speech recognition systems Park et al. (2020) by designing the training process to utilize the unlabeled data to increase the robustness of the model against noisy data Mošner et al. (2019). This technique enhances the performance of the speech recognition system in real-world situations, especially in unstructured data such as voice/imagesXie et al. (2020). The self-training process consists of multiple steps shown in **Figure 7**.

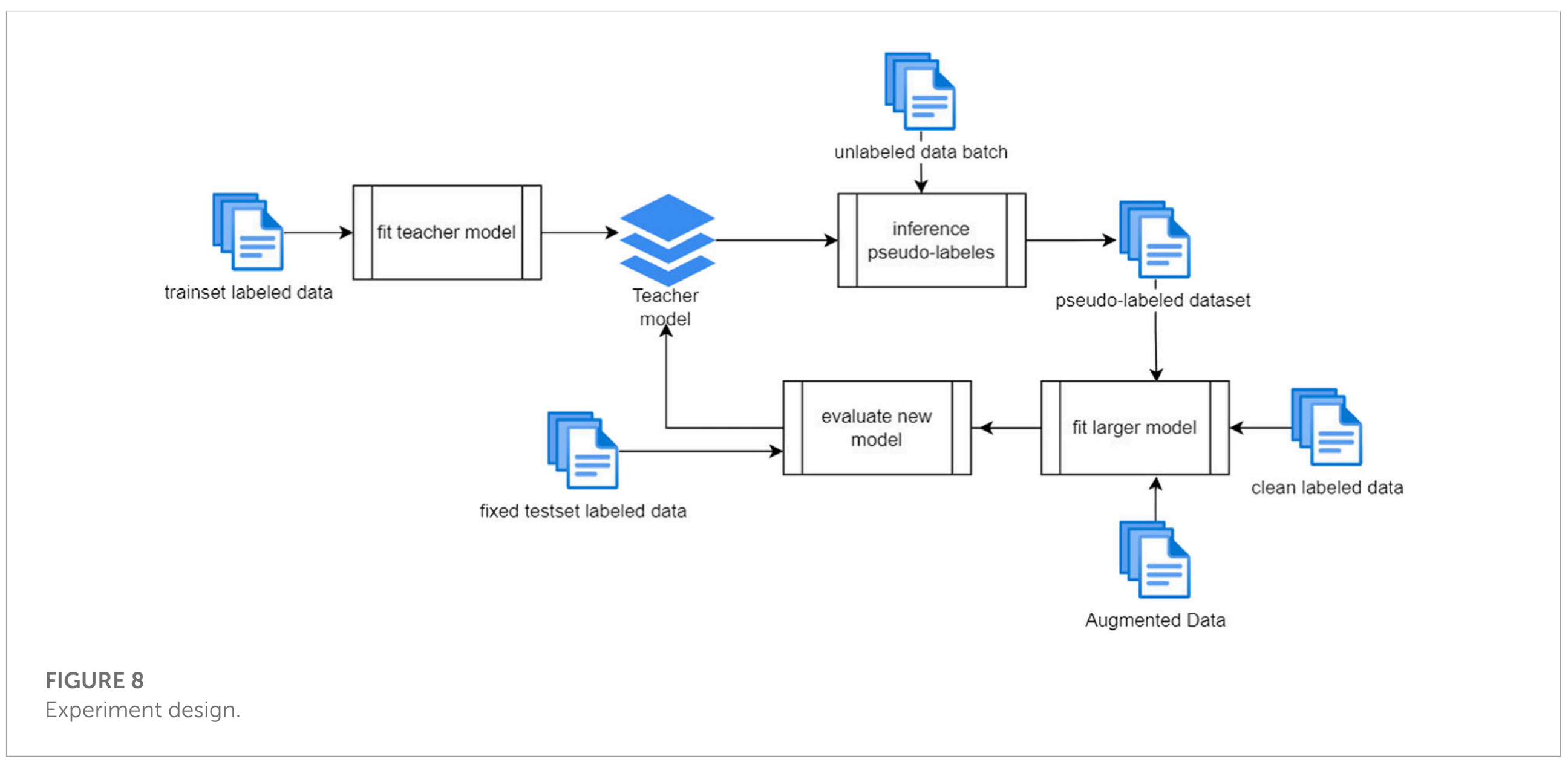

FIGURE 8
Experiment design.

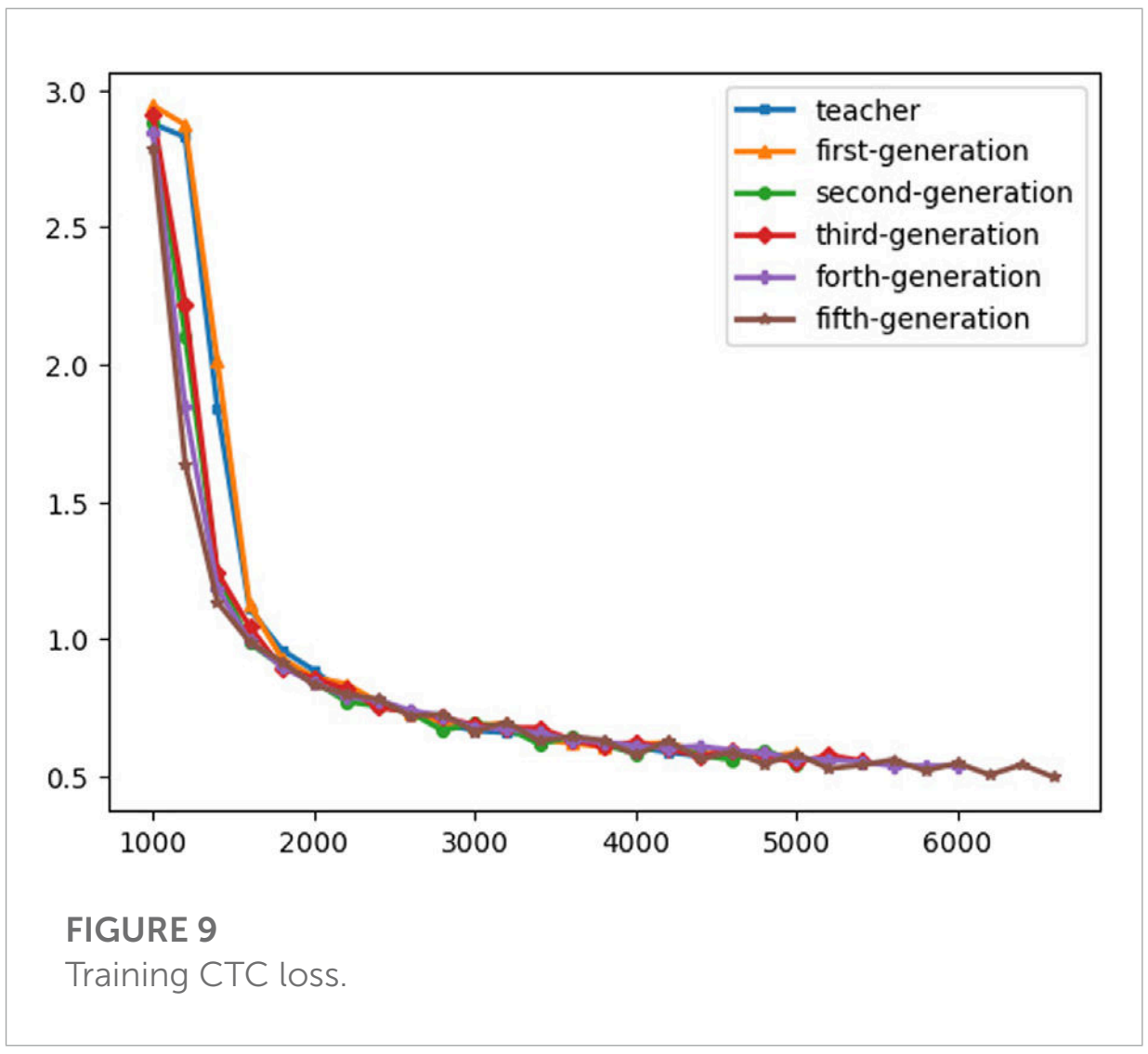

FIGURE 9
Training CTC loss.

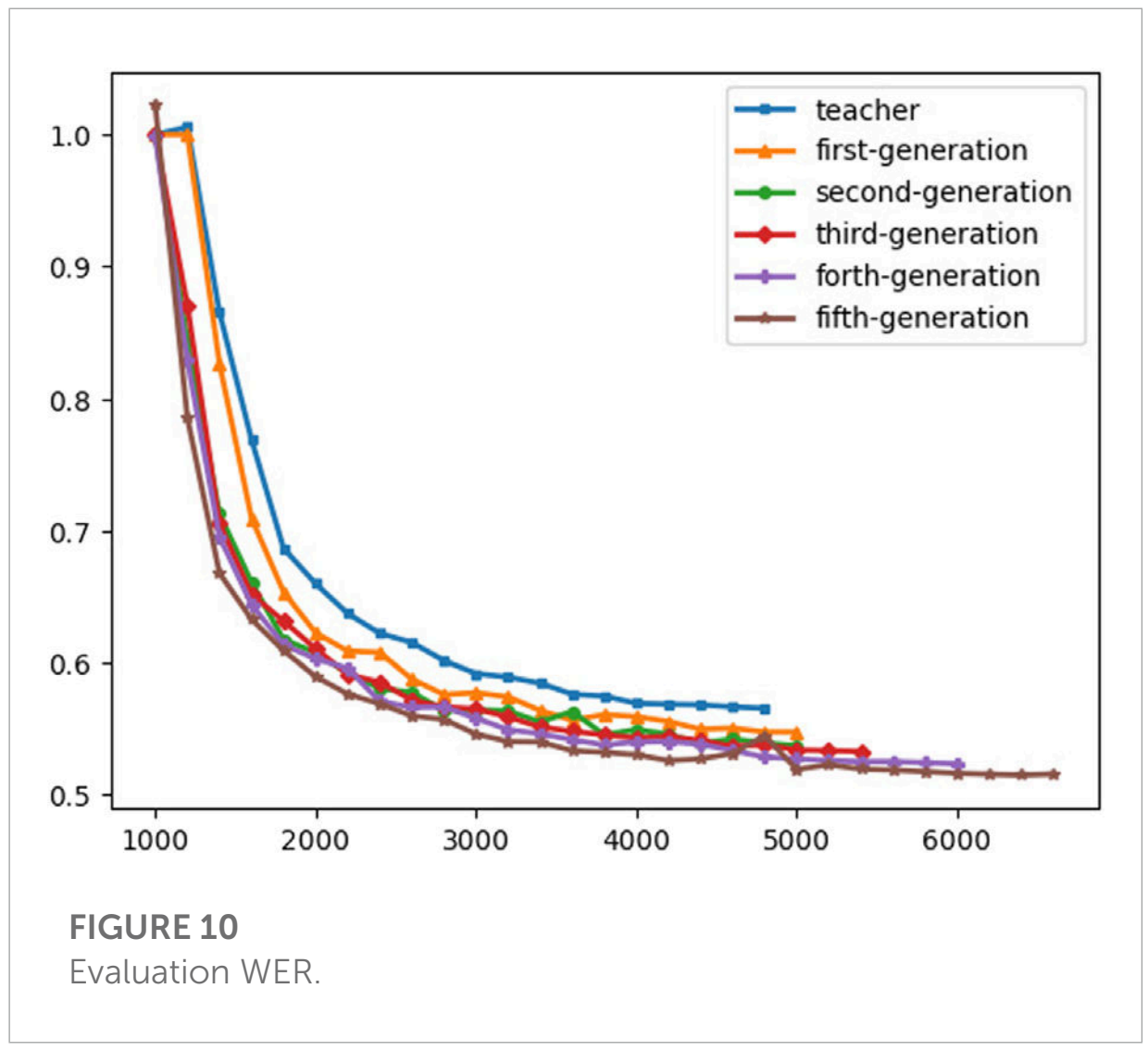

FIGURE 10
Evaluation WER.

# 8 Training

## 8.1 Experiment setup

The experiment was conducted on multi-GPU training with two NVIDIA Quadro RTX 8000-48Q, 24 vCPU, and 128 GB RAM. Fine-tuning the first teacher Wav2Vec model on the entire labeled data took 83 h.

## 8.2 Experiment design

As shown in **Figure 8**, the training process is iterative, and each iteration contains multiple steps. First of all and before starting with the training process, the data were divided into training and validation sets and the validation set holds 20% from the whole data set. The validation set is kept away from the training process, and it is used to evaluate the generalization of the teacher model. In the first step, the initial teacher model is built using the entire training data. Once the first training is completed, the resulting model is used to infer the first batch of unlabeled data with pseudo-labels. The pseudo-labeled data along with the clean labeled data and augmented data batch are used to build a larger student model and then evaluate it against the original evaluation set. The labeled data set that is used to conduct this experiment is collected from multiple Jordanian dialectal videos, and the transcripts are generated from the Google speech engine or transcripted manually from the source. Also, the whole data set is sampled into 16 kHz with 6 s as an average duration for the recordings. Our cross-lingual unsupervised trained base model is trained on 53 languages, and it can be considered SOTA in terms of phoneme error rate Conneau et al. (2020).

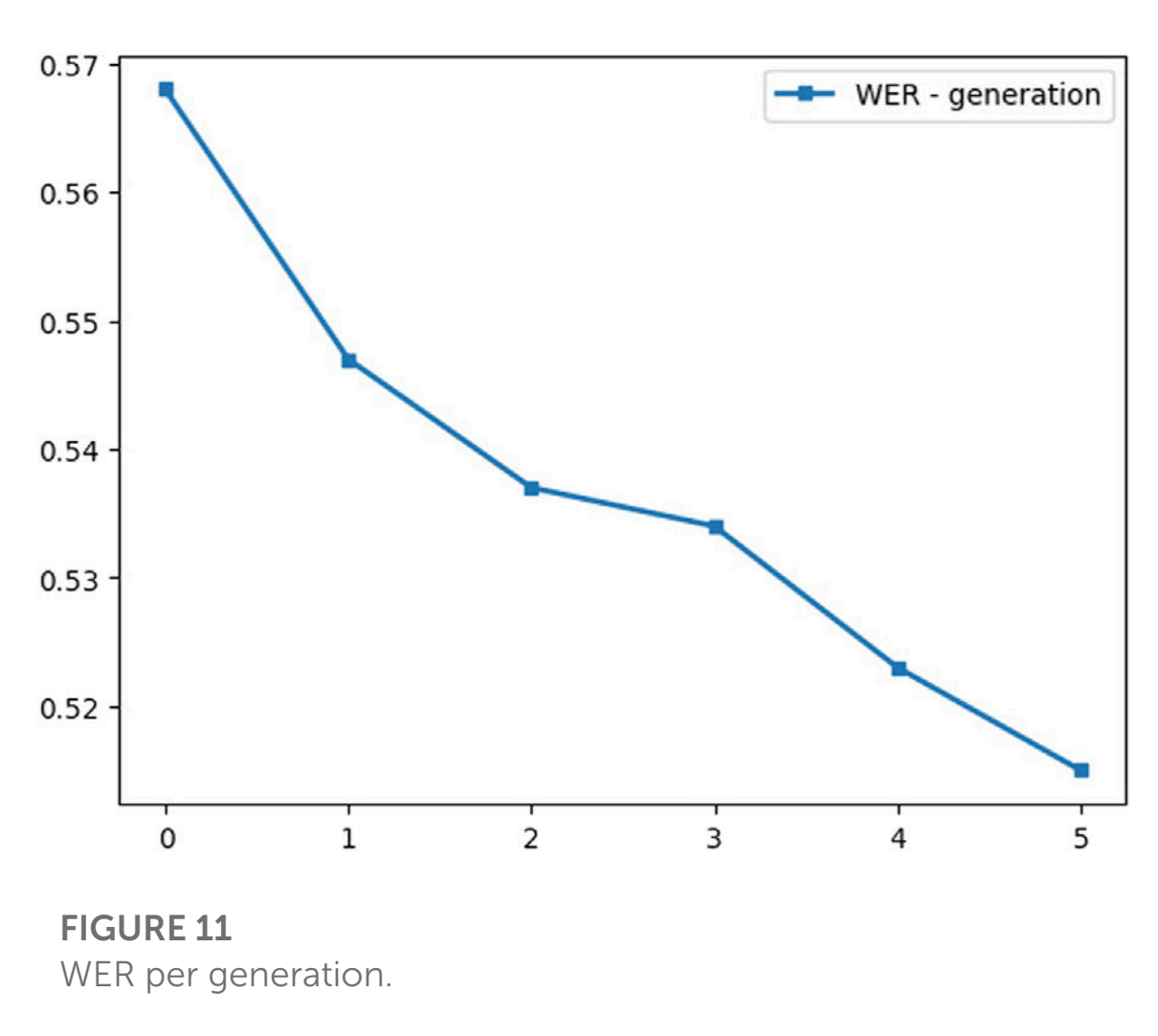

FIGURE 11
WER per generation.

## 9 Results and discussion

The experiment was conducted on a subset of the cleaned labeled data to test the training architecture against small data sets and to demonstrate the effect of the self-training process.

The labeled training data set that is used in the initial training contains 12.5 h of spoken data, and the unlabeled data are organized into five batches each batch containing 2.5 Hr. Similar to Zhang et al. (2020) that conducted a similar experiment on four generations.

Our fitness function in this research is to minimize the overall WER and to prove that utilizing the unlabeled data is important on both sides in the pretraining process and also in the post-training.

Applying noisy student training on voice data shows better generalization and rapid convergence and that is what can be concluded from **Figure 9**. In **Figure 9** applying self-training will speed up the convergence without affecting the generalization of the model.

**TABLE 4** Evaluation results.

| Model | WER | CER |
|---|---|---|
| Teacher | 0.568 | 0.279 |
| First-generation | 0.547 | 0.278 |
| Second-generation | 0.537 | 0.274 |
| Third-generation | 0.534 | 0.274 |
| Fourth-generation | 0.523 | 0.279 |
| Fifth-generation | **0.515** | **0.264** |

**Figure 11** shows the improvement in the WER at each generation, and we compare the baseline model that is trained without any unlabeled data at generation zero with the trained models on mixed labeled, unlabeled, and augmented batches. The fifth generation achieves 51% WER and shows 8% relative improvement in WER compared to the pre-trained baseline model. Moreover, utilizing the unlabeled data in the training process shows remarkable improvement in the final results of the WER score, and it's outperforming the base model by decreasing the WER by 5% and 2% in CER as shown in **Figure 10**.

Similar to Xie et al. (2020) in outperforming ImageNet, the noisy student training in voice data enhances the performance, increases the robustness, and outperforms the models that utilize the unlabeled data only for pretraining. **Table 4** shows the increase in the WER score while adding more pseudo-labeled data.

## 10 Conclusion

In this work, we investigated building a complete automatic speech recognition in low-resource languages, especially in the Jordanian dialect. This is done by ingesting the data from various sources and applying a semi-supervised approach with full utilization of the unlabeled data during the pretraining and post-training using noisy student training. Our approach outperforms the fine-tuned Wav2Vec 2.0 in terms of WER by 5.0%.

## 11 Limitations and future work

Building the Wav2Vec model on a noisy student training framework is time-consuming and needs powerful computing power in terms of GPU, RAM, and GPU memory. This is because instead of building one large model, you need to iteratively build multiple models with larger data sets and more parameters. Due to the limited resources, we cannot produce more training generations to check the ultimate improvement on the base model. In future work, different data augmentation can be compared with more noisy generations, and multiple integration methods should be explored to enable utilizing the built model in edge devices, furthermore enriching and utilizing the full data set to build a pretrained Wav2Vec model for the Arabic dialects.

## Data availability statement

The raw data supporting the conclusion of this article will be made available by the authors, without undue reservation.

## Author contributions

AS took the lead in writing the manuscript, designed the model and the computational framework, and analyzed the data. IA and RG provided critical feedback and helped shape the research, analysis, and manuscript.

## Acknowledgments

This is a short text to acknowledge the contributions of specific colleagues, institutions, or agencies that aided the efforts of the authors.

## Conflict of interest

The authors declare that the research was conducted in the absence of any commercial or financial relationships that could be construed as a potential conflict of interest.

## Publisher's note

All claims expressed in this article are solely those of the authors and do not necessarily represent those of their affiliated organizations, or those of the publisher, editors, and reviewers. Any product that may be evaluated in this article, or claim that may be made by its manufacturer, is not guaranteed or endorsed by the publisher.